\documentclass{llncs}

\usepackage[utf8]{inputenc} 
\usepackage[T1]{fontenc}    
\usepackage{hyperref}       
\usepackage{url}            
\usepackage{booktabs}       
\usepackage{amsfonts}       
\usepackage{nicefrac}       
\usepackage{microtype}      
\usepackage[dvipsnames]{xcolor}
\usepackage{graphicx}
\usepackage{verbatim}
\usepackage{array}
\usepackage{caption}
\usepackage{subcaption}
\usepackage{pgfplots}
\usepackage{pgffor}
\usepackage{tikz}
\usepackage{multirow}
\usepackage{colortbl} 
\usepackage{mathtools}
\usepackage{bm}
\usepackage{natbib}
\usepackage[inline]{enumitem}

\colorlet{vdn}{blue!70}
\colorlet{dqn}{orange!80}
\colorlet{qmix}{purple!80}
\colorlet{per}{red!70}
\colorlet{rnd}{ForestGreen}
\colorlet{3step}{Apricot}
\colorlet{5step}{Emerald}
\colorlet{7step}{MidnightBlue}
\colorlet{9step}{Lavender}

\setcitestyle{authoryear,open={[},close={]}}

\newcommand{\plotStd}[2]{%
  \addplot[color=#1, opacity=0.4, name path=f] table [x=time_step, y=#2_minus_std, col sep=comma] {datasets/#1.csv};
  \addplot[color=#1, opacity=0.4, name path=g] table [x=time_step, y=#2_plus_std, col sep=comma] {datasets/#1.csv};
  \addplot[color=#1, opacity=0.4] fill between[of=f and g];
}

\newcommand{\plotCI}[2]{%
  \addplot[color=#1, opacity=0.4, name path=f] table [x=time_step, y=#2_minus95, col sep=comma] {datasets/#1.csv};
  \addplot[color=#1, opacity=0.4, name path=g] table [x=time_step, y=#2_plus95, col sep=comma] {datasets/#1.csv};
  \addplot[color=#1, opacity=0.4] fill between[of=f and g];
}

\newcommand{\plotMean}[3][]{
    \addplot[color=#2, mark=none, line width=1pt] table [x=time_step, y=#3_mean, col sep=comma] {datasets/#2.csv};
    \ifx\relax#1\relax\else\addlegendentry{#1}\fi
}

\newcommand{\plotLevelStd}[2]{%
  \addplot[color=#1, opacity=0.4, name path=f] table [x=time_step, y=#2_minus_std, col sep=comma] {datasets/levels/#1.csv};
  \addplot[color=#1, opacity=0.4, name path=g] table [x=time_step, y=#2_plus_std, col sep=comma] {datasets/levels/#1.csv};
  \addplot[color=#1, opacity=0.4] fill between[of=f and g];
}

\newcommand{\plotLevelMean}[3][]{
    \addplot[color=#2, mark=none, line width=1pt] table [x=time_step, y=#3_mean, col sep=comma] {datasets/levels/#2.csv};
    \ifx\relax#1\relax\else\addlegendentry{#1}\fi
}

\pgfplotsset{compat=newest}
\usepgfplotslibrary{groupplots, fillbetween}

\title{Laser Learning Environment: A new environment for coordination-critical multi-agent tasks}

\author{
    Yannick Molinghen\inst{1, 2} \and
    Rahaël Avalos\inst{2} \and
    Mark Van Achter\inst{4} \and 
    Ann Nowé\inst{2} \and
    Tom Lenaerts\inst{1, 2,3}
}
\authorrunning{Molinghen et al.}
\institute{
    Machine Learning Group, Université Libre de Bruxelles, Brussels, Belgium \and
    AI Lab, Vrije Universiteit Brussel, Brussels, Belgium \and
    Center for Human-Compatible AI, UC Berkeley, USA \and 
    KU Leuven, Leuven, Belgium
}

\begin{document}
\maketitle

\begin{abstract}
    We introduce the Laser Learning Environment (LLE), a collaborative multi-agent reinforcement learning environment in which coordination is central. In LLE, agents depend on each other to make progress (\textit{interdependence}), must jointly take specific sequences of actions to succeed (\textit{perfect coordination}), and accomplishing those joint actions does not yield any intermediate reward (\textit{zero-incentive dynamics}). The challenge of such problems lies in the difficulty of escaping state space bottlenecks caused by interdependence steps since escaping those bottlenecks is not rewarded. We test multiple state-of-the-art value-based MARL algorithms against LLE and show that they consistently fail at the collaborative task because of their inability to escape state space bottlenecks, even though they successfully achieve perfect coordination. We show that $Q$-learning extensions such as prioritised experience replay and $n$-steps return hinder exploration in environments with zero-incentive dynamics, and find that intrinsic curiosity with random network distillation is not sufficient to escape those bottlenecks. We demonstrate the need for novel methods to solve this problem and the relevance of LLE as cooperative MARL benchmark.
\end{abstract}

\section{Introduction}

Many problems, ranging from societal to technological, are inherently multi-agent and often require coordination (or cooperation) among the agents to achieve individually and collectively defined goals \citep{cao_2013_autonomous, klima_2018_space}. While evolution has provided humans with the skills to deal with such tasks, researchers and engineers have to train the artificial agents in a much shorter time span to also manage such tasks. 

We are interested in Reinforcement Learning \citep[RL]{sutton_barto_2018_rlbook} as one of the branches of Machine Learning that holds the promise of training such agents by interacting with their environment. Deep RL has made dazzling progress in recent years with deep $Q$-learning, showing human or superior to human performance in a wide range of single-agent situations \citep{dqn_mnih_2015}, and this progress influenced Multi-Agent Reinforcement Learning \citep[MARL]{panait_cooperative_2005_marl}.

In comparison to single-agent RL, centralised MARL faces the issue of the exponential growth of state and joint action spaces with the number of agents, making this approach intractable even for relatively small problems. Decentralised MARL approaches avoid the exponential growth of the action space at the cost of nonstationarity \cite{laurent_world_2011_non-markovian}: Since multiple learning agents adapt their policy over time, each agent continuously has to adapt to the changing policy of the other agents, making it more challenging to acquire robust and general policies. To mitigate that effect, \citet{oliehoek_optimal_2008_ctde} introduce the paradigm of Centralised Training with Decentralised Execution (CTDE) that has demonstrated how successful it can be in complex cooperative multi-agent tasks \citep{vdn_sunehag_value-decomposition_2018, rashid_qmix_2018, Avalos2022LocalLearningAAMAS}.

In the last five years, a variety of environments have been developed for cooperative MARL. Among others, the Multi-agent Particle Environment \cite[MPE]{mpe_lowe2017multi, mpe_mordatch2017emergence}, the StarCraft Multi-Agent Challenge \cite[SMAC]{samvelyan19smac}, the Hanabi Learning Environment \cite[HLE]{bard_hanabi_2020} and Overcooked \cite{overcooked_wu_wang2021too} respectively aim at studying different aspects of cooperative multi-agent problem-solving such as the ability to infer the intentions of other agents or the emergence of basic compositional language.

\paragraph{Contributions} In this work, we focus on fully cooperative multi-agent problems where agents optimise a single shared reward. We identify a category of such problems that is not well studied, introduce the Laser Learning Environment (LLE) that fits in that category and test state-of-the-art CTDE methods against it. To the best of our knowledge, LLE exhibits a unique combination of three properties: 
\begin{enumerate*}[label=\arabic*)]
    \item \textit{perfect coordination}: failing to coordinate can be fatal;
    \item \textit{interdependence}: agents need each other to progress;
    \item \textit{zero-incentive dynamics}: key steps toward success are not rewarded.
\end{enumerate*}
We then show how agents successfully achieve perfect coordination but are unable to overcome the zero-incentive dynamics and escape state space bottlenecks, even when using $Q$-learning extensions such as prioritised experience replay \citep{schaul_prioritized_2016_per} and $n$-steps return \citep{Watkins_1989_n-step-return}. Finally, we show that intrinsic curiosity with Random Network Distillation \citep{burda_exploration_2018_rnd} does not overcome the state space bottlenecks created by the interdependence of the agents. Together, our results demonstrate that LLE is a relevant benchmark for future work and aim to point researchers in new relevant directions of cooperative MARL research.

\section{Background}

\subsection{Multi-agent Markov Decision Process}
\label{sec:mdp}
A Multi-agent Markov Decision Process \citep[MMDP]{boutilier_planning_1996} is described as a tuple $\left<n, S, A, T, R, s_0, s_f, \gamma\right>$ where $n$ is the number of agents, $S$ is the set of states, $A \equiv A_1 \times \dots \times A_n$ is the set of joint actions and $A_i$ is the set of actions of agent $i$, $T\colon S \times A \rightarrow \Delta_S$ is a function that gives the probability of transitioning from state $s$ to state $s'$ by taking action $a$, $R: S \times A \times S \rightarrow \mathbb{R}$ is the function that gives the reward obtained by transitioning from $s$ to $s'$ by performing joint action $\bm{a}$, $s_0 \in S$ is the initial state, $s_f$ is the final state and $\gamma \in \left[0, 1\right)$ is a discount factor.\\
A transition is defined as $\tau = \left<s, \bm{a}, r, s'\right>$ with $ s, s' \in S, \bm{a} \in A, r \in \mathbb{R}$. An episode of length $l$ is a sequence of transitions $\tau_1, \dots, \tau_l$ such that $\tau_1 = \left< s_0, \bm{a}, r, s'\right>$ and $\tau_l = \left<s_{l-1}, \bm{a}, r, s_f\right>$. Each agent $i$ acts according to a policy $\pi_i \colon S  \rightarrow \Delta_{A_i}$ and their objective is to find the joint policy $\bm{\pi} = \left<\pi_1, \dots, \pi_n\right>$ that maximises their expected discounted reward $\mathbb{E}\left[\sum_{t=0}^{\infty} \gamma^{t} R_t | s=s_0, \bm{\pi}\right]$. The action-value function of a policy measures the expected return obtained by taking action $\bm{a}$ is state $s$ by following policy $\bm{\pi}$. In particular, we define the joint action-value function $Q^{\bm{\pi}}: S \times \bm{A} \rightarrow \mathbb{R}$.\\
Scenarios where agents receive individual observations of the state instead of the full state are referred to as Decentralised Partially Observable Markov Decision Processes \citep{oliehoek_concise_2016_pomdp}.

\subsection{\textit{Q}-value factorisation}
\label{sec:ctde}

\citet{laurent_world_2011_non-markovian} showed that multi-agent systems suffer from a non-stationarity problem (\citet{tuyls_multiagent_2012_moving_target} also refer to it as the multi-agent moving target problem) because learning agents perceive the other learning agents as part of the environment. \citet{claus_dynamics_1998} have shown that naive implementations of Independent Q-Learning (IQL) were often unsuccessful, even for very simple tasks, partly because of this non-stationarity.

To tackle this non-stationarity problem, \citet{vdn_sunehag_value-decomposition_2018} introduce Value Decomposition Network (VDN), an algorithm based on the concept of $Q$-value factorisation \citep{oliehoek_exploiting_2008_factorisation} in which each agent $i$ has its own utility function $Q_i: S \times A_i \rightarrow \mathbb{R}$. VDN decomposes the joint $Q$-value into a simple sum of the agents' utility.

This factorisation allows for decentralised execution thanks to the Individual Global Max property \citep[IGM]{son_qtran_2019} that ensures consistency in action selection between the centralised training and the decentralised execution.
QMIX \citep{rashid_qmix_2018} extends VDN by allowing more complex factorisation using a monotonically increasing hype-network conditioned on the state. 



\subsection{Cooperative multi-agent environments}
\label{sec:environemnts}

In the last few years, several cooperative multi-agent environments have emerged to study different aspects of the cooperative multi-agent problem and we give here an overview of popular environments with discrete action spaces.

\subsubsection{The StarCraft Multi-Agent Challenge} \citep[SMAC]{samvelyan19smac} comes with a wide range of maps and offers a complex cooperative partially observable problem where individual agents of the same team have to defeat the opponent team in a short skirmish. This environment has been introduced to study whether agents could learn complex behaviours such as kiting.

\subsubsection{Overcooked} \citep{overcooked_wu_wang2021too} is a cooperative environment in which multiple agents receive recipes and have to cook them before serving them on a plate. On top of spatial movements, this environment has been introduced to study the ability of agents to infer the hidden intentions of others as well as the ability of agents to learn when to split the tasks amongst themselves (divide and conquer) and when to work on the same task (cooperate).

\subsubsection{Hanabi Learning Environment} \citep[HLE]{bard_hanabi_2020} is an environment that comes from a board game with the same name. In Hanabi, every player plays one after the other and has to decide whether to play a card or give a clue to a teammate. This environment is well suited to study reasoning, beliefs and intentions of other players.

\subsubsection{The Multi-agent Particle Environment} \citep[MPE]{mpe_lowe2017multi, mpe_mordatch2017emergence} is a set of cooperative and competitive 2D tasks with a dense reward signal. It has both partially and fully observable variants and offers scenarios that involve communication. This environment has been introduced to study the emergence of a basic compositional language.

\section{The Laser Learning Environment}
\label{sec:LLE}

\begin{figure}
  \centering
  \scriptsize
  \begin{tikzpicture}
    \node[anchor=south west, inner sep=0] (image) at (0,0) {\includegraphics[width=0.4\textwidth]{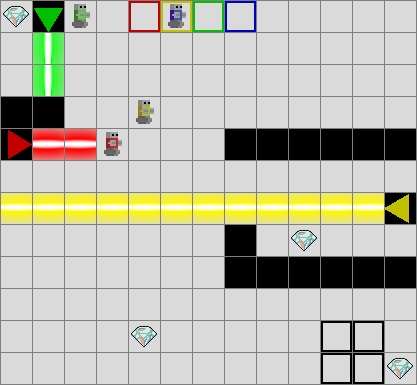}};
    
     \draw[black,thick] (1.48, 4.08) rectangle ++ (1.6, 0.5) node (start) {};
    \draw[black,thick] (-0.03, 1.82) rectangle ++ (4.58, 0.5) node (laser) {};
    \draw[black,thick] (3.72, -0.03) rectangle ++ (0.84, 0.84) node (exit) {};
    
    \node[draw,fill=white, anchor=west] at (3.2, 4.3) {Start tiles};
    \node[draw,fill=white, anchor=west] at (0.5, 1.55) {Yellow laser};
    \node[draw,fill=white, anchor=west, align=center] at (5, 2.1) {Laser\\ source};
    \node[draw,fill=white, anchor=west] at (1.3, 0.2) {Gem};
    \node[draw,fill=white, anchor=west] at (1.9, 3.2) {Yellow agent};
    \node[draw,fill=white, anchor=east] at (3.7, 0.4) {Exit tiles};
  \end{tikzpicture}
  \caption{Level 6 of LLE, which has 4 agents, 3 lasers and 4 gems. Agent red blocks the red laser, making it possible for the other agents to pass to the lower part of the grid world. Additional blocking of the yellow laser is required for them to all pass and reach the exit tiles.}
    \label{fig:lvl6-annotated}
\end{figure}
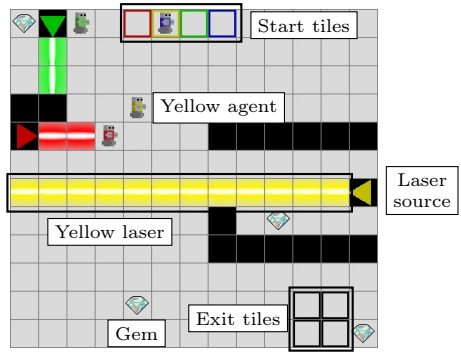

We introduce here the Laser Learning Environment (LLE), a multi-agent grid world populated by agents of different colours (see \autoref{fig:lvl6-annotated}) and with different types of tiles: floor, start, wall, laser, laser source, gem, and exit tiles. The main objective in LLE is for each agent to reach an exit tile, while additional points can be gathered by collecting gems along the way. The game is cooperative since agents must help each other to pass laser beams:  if an agent's colour doesn't match the colour of a laser beam, this agent dies and the game ends. However, when an agent of matching colour stands in line with a beam, it blocks the laser and allows other agents to reach areas of the map that they could not access by themselves.

LLE comes with 6 different maps (Appendix~\ref{apx:maps}) and can also generate random solvable maps of arbitrary sizes based on multiple parameters (number of agents, number of gems, number of lasers and wall density). Randomly generated maps can also be constrained to some degree of difficulty depending on how many steps of \textit{perfect coordination} (see \autoref{sec:motivations}) are required to complete the level. While future work will address the issue of curriculum learning \citep{parker2022evolving}, in this work, we focus on the configuration visualised in  \autoref{fig:lvl6-annotated}.

\subsection{Motivations}
\label{sec:motivations}
As explained in \autoref{sec:environemnts}, there already exist several multi-agent cooperative environments that are suitable for studying various fields of MARL. However, LLE aims at studying a new range of problems. As far as we were able to find the information, LLE introduces new complexities in the study of cooperative problem solving due to a combination of three properties: \textit{perfect coordination}, \textit{interdependence} and \textit{zero incentive dynamics}.

\subsubsection{Perfect coordination}~
is defined as the property of a policy in which agents simultaneously take a specific sequence of actions such that any agent that singlehandedly deviates from this policy in state $s_t$ would directly lead to a penalty in state $s_{t+1}$ and possibly the early termination of the game.

Unlike all the environments presented in \autoref{sec:environemnts}, LLE requires agents to achieve perfect coordination when blocking lasers. Focussing on agents red and yellow in \autoref{fig:lvl6-annotated}, it requires perfect coordination for agent yellow to cross the red laser: if agent yellow goes down and agent red releases the laser, agent yellow would die and the game would terminate with a punishment.
Although Hanabi has the comparable property of directly punishing agents that fail to coordinate, the environment is such that agents play one after the other. For that reason, Hanabi does not require \textit{perfect coordination}.

\subsubsection{Interdependence} \label{sec:prop-interdependence}~ 
expresses how much a set of agents relies on another set of agents to perform a particular sequence of actions in order to make progress in the collective task.

Interdependence introduces \emph{bottlenecks in the state space} in the steps that require coordination. A high level of interdependence means that from the start state, the end state that maximises the collective expected discounted return is located behind such interdependence bottlenecks. Inversely, a low level of interdependence means that agents can explore most of the state space without relying on each other.

The maps presented in Appendix~\ref{apx:maps} have been designed with increasing levels of interdependence in mind, with level 6 having the highest level of interdependence. In comparison, SMAC, MPE and HLE challenge agents with situations where any agent can explore the state space, sometimes even finish the game, regardless of their teammates' actions.

Overcooked explicitly uses the concept of sub-tasks in the form of recipes that also enclose the reachable state space until the current sub-task is solved. However, agents can most of the time accomplish those sub-tasks by themselves and are therefore not interdependent. Arguably, there is \textit{interdependence} in completely split maps where agents do not have access to every kitchen tool (e.g.: full-divider map) and must pass items over the counter to the other to complete recipes.

\subsubsection{Zero-incentive dynamics} defines the property of an environment in which overcoming bottlenecks in the state space is not rewarded. Intuitively, an environment with zero-incentive dynamics does not reward agents for succeeding at key (cooperative) dynamics. 

As detailed in \autoref{sec:rewards}, blocking lasers in LLE does not provide any reward, and laser-blocking are state space bottlenecks caused by interdependence. Consequently, LLE has zero-incentive dynamics. Overcooked on the other hand provides rewards for finishing recipes and therefore does not have zero-incentive dynamics, even in the full-divider map that does have interdependence. We discuss in \autoref{sec:results} that the combination of interdependence and zero-incentive dynamics makes it difficult for agents to escape regions surrounded by lasers during the training process, which makes LLE a challenging exploration task.

We distinguish zero-incentive dynamics from temporal credit assignment \citep{sutton_1984_credit_assignment}, which is related to the ability of a policy to determine which action takes credit for some later reward, and not a property of the environment.

\subsection{States}
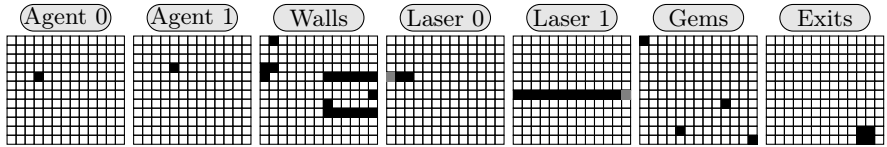
\begin{figure}[t]
    \centering
    \begin{tikzpicture}[scale=0.12]
        \filldraw[fill=gray!20, rounded corners=5pt] (1.5,15.5) rectangle ++(10,-2.95) node[midway,align=center]{Agent 0};
        \foreach \x in {0,...,12}{
            \foreach \y in {0,...,11}{
                \draw (\x,\y) rectangle ++(1,1);
            }
        }
        \fill[black] (3,7) rectangle ++(1, 1);
    \end{tikzpicture} 
    \begin{tikzpicture}[scale=0.12]
        \filldraw[fill=gray!20, rounded corners=5pt] (1.5,15.5) rectangle ++(10,-2.95) node[midway,align=center]{Agent 1};
        \foreach \x in {0,...,12}{
            \foreach \y in {0,...,11}{
                \draw (\x,\y) rectangle ++(1,1);
            }
        }
        \fill[black] (4,8) rectangle ++(1, 1);
    \end{tikzpicture}
    \begin{tikzpicture}[scale=0.12]
        \filldraw[fill=gray!20, rounded corners=5pt] (1.5,15.5) rectangle ++(10,-2.95) node[midway,align=center]{Walls};
        \foreach \x in {0,...,12}{
            \foreach \y in {0,...,11}{
                \draw (\x,\y) rectangle ++(1,1);
            }
        }
        \fill[black] (0,7) rectangle ++(1, 1);
        \fill[black] (0,8) rectangle ++(1, 1);
        \fill[black] (1,8) rectangle ++(1, 1);
        \fill[black] (1,11) rectangle ++(1, 1);

        \foreach \x in {7,...,12}{
            \fill[black] (\x,7) rectangle ++(1, 1);
        }

        \foreach \x in {7,...,12}{
            \fill[black] (\x,3) rectangle ++(1, 1);
        }
        \fill[black] (7,4) rectangle ++(1, 1);
        \fill[black] (12,5) rectangle ++(1, 1);
    \end{tikzpicture}
    \begin{tikzpicture}[scale=0.12]
        \filldraw[fill=gray!20, rounded corners=5pt] (1.5,15.5) rectangle ++(10,-2.95) node[midway,align=center]{Laser 0};
        \foreach \x in {0,...,12}{
            \foreach \y in {0,...,11}{
                \draw (\x,\y) rectangle ++(1,1);
            }
        }
        \fill[gray] (0,7) rectangle ++(1, 1);
        \fill[black] (1,7) rectangle ++(1, 1);
        \fill[black] (2,7) rectangle ++(1, 1);
    \end{tikzpicture}
    \begin{tikzpicture}[scale=0.12]
        \filldraw[fill=gray!20, rounded corners=5pt] (1.5,15.5) rectangle ++(10,-2.95) node[midway,align=center]{Laser 1};
        \foreach \x in {0,...,12}{
            \foreach \y in {0,...,11}{
                \draw (\x,\y) rectangle ++(1,1);
            }
        }
        \foreach \x in {0,...,11}{
            \fill[black] (\x,5) rectangle ++(1, 1);
        }
        \fill[gray] (12,5) rectangle ++(1, 1);
    \end{tikzpicture}
    \begin{tikzpicture}[scale=0.12]
        \filldraw[fill=gray!20, rounded corners=5pt] (1.5,15.5) rectangle ++(10,-2.95) node[midway,align=center]{Gems};
        \foreach \x in {0,...,12}{
            \foreach \y in {0,...,11}{
                \draw (\x,\y) rectangle ++(1,1);
            }
        }
        \fill[black] (0,11) rectangle ++(1, 1);
        \fill[black] (12,0) rectangle ++(1, 1);
        \fill[black] (4,1) rectangle ++(1, 1);
        \fill[black] (9,4) rectangle ++(1, 1);
    \end{tikzpicture}
    \begin{tikzpicture}[scale=0.12]
        \filldraw[fill=gray!20, rounded corners=5pt] (1.5,15.5) rectangle ++(10,-2.95) node[midway,align=center]{Exits};
        \foreach \x in {0,...,12}{
            \foreach \y in {0,...,11}{
                \draw (\x,\y) rectangle ++(1,1);
            }
        }
        \foreach \x in {10,...,11}{
            \foreach \y in {0,...,1}{
                \fill[black] (\x,\y) rectangle ++(1, 1);
            }
        }
    \end{tikzpicture}
    \caption{Representation of the state shown in \autoref{fig:lvl6-annotated}. The layers ``Agent 2'', ``Agent 3'', ``Laser 2'' and ``Laser 3'' were omitted for the sake of conciseness. Each layer encodes the location of a specific type of object of the grid world (walls, agents' locations, \dots). White squares represent $0$s, black squares are $1$s and grey squares are $-1$.}
    \label{fig:layered}
\end{figure}

The state of LLE encodes the location of grid-world elements layer by layer as illustrated in \autoref{fig:layered}. These items are namely agents, walls, lasers, laser sources, gems and exits locations. There is one layer per agent and one layer per laser colour. 
This representation has the advantage of being very generic as long as the size of the map and the number of agents remain identical, allowing future work in the field of generalisation and curriculum learning \citep{parker2022evolving}.

\subsection{Actions}
The action space of LLE is discrete and the possible actions are \texttt{NORTH}, \texttt{EAST}, \texttt{SOUTH}, \texttt{WEST} and \texttt{STAY}, which is required for laser-blocking purposes. Actions are identical for all agents.

LLE prevents agents from entering invalid tiles by providing the set of available actions for each agent at each time step. An agent cannot enter walls, move beyond the grid boundaries or enter a tile that is currently occupied by another agent. This prevents multiple types of conflicts that can occur in grid world problems such as edge, following and swapping conflicts \citep{stern_multi-agent_2017_conflicts}.  Additionally, once an agent enters an exit tile, it cannot leave it anymore and the only action allowed is \texttt{STAY}. 

Vertex conflicts -- when two or more agents enter the same tile -- can unfortunately not be prevented with information on action availability. As a result, when agents provoke a vertex conflict by trying to move to the same tile, their actions are replaced by \texttt{STAY}.

\subsection{Rewards}
\label{sec:rewards}
The reward function of LLE has a single scalar output as the team reward, which is the result of the joint action of the agents at a given time step. Collecting a gem or entering an exit tile provides a collective reward of $+1$. Finishing the game, i.e. when all agents are on an exit tile, also provides an additional reward of $+1$. However, if any agent dies at a time step, the episode ends and the reward is set to $-1$ times the number of agents that have died.

\subsection{Metrics}
\label{sec:metrics}
Two metrics are used here to evaluate the performance of the LLE agents: the score and the exit rate.

\subsubsection{Score}
The score refers to the undiscounted sum of rewards throughout an episode. This is metric is close to the one the agents are trained to maximise, i.e. the \textit{discounted} sum of rewards over the course of an episode, but does not give insight on the time taken to achieve the task. The maximal score of an LLE environment can be calculated as follows: given a map with $n$ agents and $g$ gems, we can compute the maximum score as $\text{maximum score} = n + g + 1$.

\subsubsection{Exit rate}
The exit rate is defined as the proportion of agents that reach an exit tile at the end of an episode. Since the objective in every LLE is to exit the level, this metric gives an estimation of how close agents are to the objective. An exit rate of $1$ means that all agents have successfully exited the level.\\

The combination of those two metrics gives insight into the agents' behaviour. As long as the exit rate is below 1, it means that the agents are not able to finish the level. If the exit rate is 1, then analysing the score allows us to know if agents have collected all the gems.

\subsection{Implementation}
LLE is implemented in Rust which makes it an extremely fast environment. LLE comes with a strongly type hinted Python interface for seamless integration with common MARL libraries and is extensively tested in both Python and Rust.

\section{Experiments}
\label{sec:experients}

We analyse the score and exit rate on level 6\footnote{Results for each level can be found in Appendix~\ref{apx:all-results}.} shown in \autoref{fig:lvl6-annotated} with Independent deep $Q$-Learning \citep[IQL]{dqn_mnih_2015}, Value Decomposition Network \citep[VDN]{vdn_sunehag_value-decomposition_2018} and QMIX \citep{rashid_qmix_2018}. Then, we discuss in \autoref{sec:discussion} the results with regard to  perfect coordination, interdependence and zero-incentive dynamics.

In their respective papers, VDN and QMIX have been introduced in the scope of partially observable environments while LLE is fully observable. We treat the fully observable case as the particular case of Dec-POMDP where the individual observations are equal to the state. We motivate the usage of QMIX by the fact that it has demonstrated its superior representational capability in comparison to VDN with a fully observable example referred to as \textit{Two-Step game} \citep{rashid_qmix_2018}. Therefore, we use a feed-forward neural network instead of a recurrent one.

\paragraph{Experimental setup} In our experiments, agents interact with the environment for one million time steps and policies are updated every 5 steps on a batch of 64 transitions. Agents use an $\epsilon$-greedy policy linearly annealed from $1$ to $0.05$ over 500k time steps and use a replay memory of 50k transitions. We use Double $Q$-learning for all algorithms. The target network is updated via soft updates with $\tau=0.01$ \citep{ddpg_2016} which has experimentally provided better results in our experiments than periodic hard updates of the target network. All hyperparameters can be found in \autoref{apx:hyperparameters}.

The utility values $Q_i$ of each agent $i \in \left\{1, \dots, n\right\}$ are estimated with a convolutional neural network \citep{lecun_gradient-based_1998_cnn} to take advantage of the spatial nature of the layered observations. Since the observations are designed such that every \textit{pixel} gives information, the stride of the convolutions is $1$. \autoref{apx:nn_architecture} provides a more detailed description of the neural network architecture. Since LLE is fully observable and agents share the same neural network weights, we concatenate the flattened output of the CNN with a one-hot encoding of the agent id. This allows the neural network to output different $Q_i$-values for each agent.

We define the maximal episode duration to be $\lfloor{\frac{width \times height}{2}\rceil}$ steps after which the episode is truncated and taken to an end. This heuristic for episode duration allows the agents to discover a lot of dynamics of the environment without polluting the replay memory too much with useless transitions. For instance in level 6, if agents red and yellow reach the exit tiles but did not open the way for blue and green, then the remaining time steps of the episode would just be the latter two waiting behind the beam.

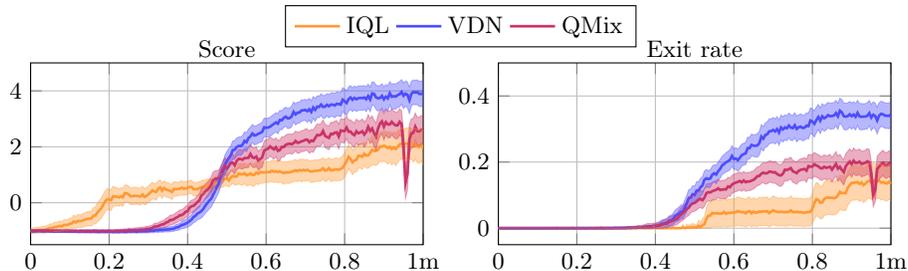
\begin{figure}[t]
    \centering
    \begin{tikzpicture}
        \begin{groupplot}[
            group style={group size=2 by 1, horizontal sep=1cm},
            height=4cm,
            width=6.8cm, 
            grid=major,
            xtick={0, 200000, 400000, 600000, 800000, 1000000},
            xticklabels={0, 0.2, 0.4, 0.6, 0.8, 1m},
            scaled x ticks=false,
            legend columns=-1,
            title style={yshift=-0.25cm},
        ]
            \nextgroupplot[title={Score}, legend to name=zelda, ymin=-1.5, ymax=5, xmin=0, xmax=1000000]            
                \plotMean[IQL]{dqn}{score};
                \plotMean[VDN]{vdn}{score};
                \plotMean[QMix]{qmix}{score};

                \plotCI{dqn}{score};
                \plotCI{qmix}{score};
                \plotCI{vdn}{score};
                                        
                \coordinate (top-left) at (rel axis cs:0,1);
                \coordinate (bot-left) at (rel axis cs:0,0);
                
            \nextgroupplot[ymin=-0.05, ymax=0.5, xmin=0, xmax=1000000, title={Exit rate}]
                \plotMean{dqn}{exit_rate};
                \plotMean{vdn}{exit_rate};
                \plotMean{qmix}{exit_rate};
            
                \plotCI{dqn}{exit_rate};
                \plotCI{qmix}{exit_rate};
                \plotCI{vdn}{exit_rate};

                \coordinate (right) at (rel axis cs:1,1);
                \coordinate (bot) at (rel axis cs:1,0);
        \end{groupplot}
        \path (top-left)--(bot) coordinate[midway] (center);
        \path (top-left)--(bot-left) coordinate[midway] (center-left);
        \path (top-left)--(right) coordinate[midway] (h-center);
        \node[above=0.2cm, inner sep=0pt] at (h-center) {\pgfplotslegendfromname{zelda}};
    \end{tikzpicture}
    \caption{Training score and exit rate over time for IQL, VDN and QMIX on level 6 (\autoref{fig:lvl6-annotated}). The maximal achievable score is 9. Results averaged on 20 different seeds and shown with 95\% confidence interval, capped by the minimum and maximum.}
    \label{fig:dqn-vdn-qmix}
\end{figure}

\paragraph{Foreanalysis}
Level 6 (\autoref{fig:lvl6-annotated}) is of size $12 \times 13$ and has 4 agents and 4 gems. The maximal score is hence $4+4+1=9$ as explained in \autoref{sec:metrics}. The optimal policy in level 6 is the following: 
\begin{enumerate*}[label=\textbf{\roman*)}]
    \item Agent green should collect the gem in the top left corner;
    \item Agent red should block the red laser and wait for every other agent to cross;
    \item Agent yellow should cross the red laser and collect the gem that only he can collect near the yellow source;
    \item Agent yellow should block the laser for every agent to cross;
    \item Agents should collect the remaining gems on the bottom half;
    \item Agents should go to the exit tiles.
\end{enumerate*}
The length of such an episode is $\approx 30$ time steps, well below the time limit of $\left\lfloor\frac{12 \times 13}{2}\right\rceil = 78$ steps.

\subsection{Baseline results}
\label{sec:results}

\autoref{fig:dqn-vdn-qmix} shows the mean score and exit rate over the course of training on level 6 (\autoref{fig:lvl6-annotated}). VDN performs best on this map. That being said, none of the algorithms ever reaches the highest possible score of $9$ and at most half of the agents ever reach the end exit tiles.

Looking into the results, the best policy learned only completes items i), iii) and v). Agents red and yellow escape the top half of the map, collect gems on that side and reach the exit, while agents green and blue are not waited for. This policy yields a score of 6 and an exit rate of 0.5. We could introduce reward shaping in order to drive the agents towards a better solution more easily. However, reward shaping is notoriously difficult to achieve properly and can drive agents towards unexpected (and likely undesired) behaviours \citep{amodei_2016_reward}.

\subsection{Results with \textit{Q}-learning extensions}
\label{sec:results-extensions}

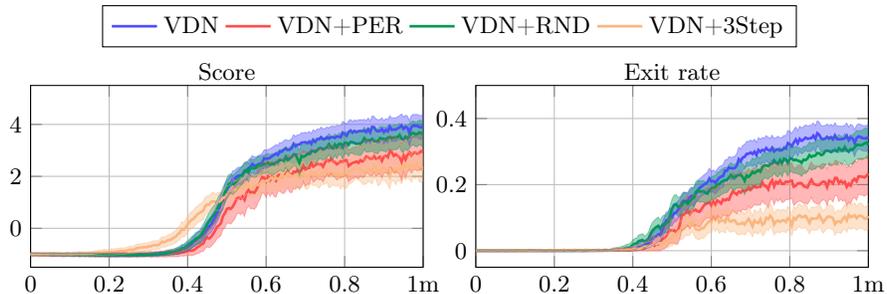
\begin{figure}[t]
    \centering
    \begin{tikzpicture}
        \begin{groupplot}[
            group style={group size= 3 by 1, horizontal sep=1cm},
            height=4cm,
            width=6.8cm, 
            grid=major,
            legend columns=-1,
            xtick={0, 200000, 400000, 600000, 800000, 1000000},
            xticklabels={0, 0.2, 0.4, 0.6, 0.8, 1m},
            scaled x ticks=false,
            title style={yshift=-0.25cm},
        ]
            \nextgroupplot[title={Score}, legend to name=link, ymin=-1.5, ymax=5.5, xmin=0, xmax=1000000]
                \plotMean[VDN]{vdn}{score};
                \plotMean[VDN+PER]{per}{score};
                \plotMean[VDN+RND]{rnd}{score};
                \plotMean[VDN+3Step]{3step}{score};
            
                \plotCI{vdn}{score};
                \plotCI{per}{score};
                \plotCI{rnd}{score};
                \plotCI{3step}{score};
                        
                \coordinate (top-left) at (rel axis cs:0,1);
                \coordinate (bot-left) at (rel axis cs:0,0);
                
            \nextgroupplot[title={Exit rate}, ymin=-0.05, ymax=0.5, xmin=0, xmax=1000000, xshift=-0.3cm]
                \plotMean{vdn}{exit_rate};
                \plotMean{per}{exit_rate};
                \plotMean{rnd}{exit_rate};
                \plotMean{3step}{exit_rate};
            
                \plotCI{vdn}{exit_rate};
                \plotCI{per}{exit_rate};
                \plotCI{rnd}{exit_rate};
                \plotCI{3step}{exit_rate};
                                
                \coordinate (right) at (rel axis cs:1,1);
                \coordinate (bot) at (rel axis cs:1,0);
        \end{groupplot}
        \path (top-left)--(bot) coordinate[midway] (center);
        \path (top-left)--(bot-left) coordinate[midway] (center-left);
        \path (top-left)--(right) coordinate[midway] (h-center);
        \node[above=0.5cm, inner sep=0pt] at (h-center) {\pgfplotslegendfromname{link}};
    \end{tikzpicture}
    \caption{Training score and exit rate over training time for VDN, VDN with PER, VDN with RND and VDN with 3-step return on level 6. The maximal score that agents can reach on level 6 of an episode is $9$. Results are averaged on 20 different seeds and shown with 95\% confidence intervals}
    \label{fig:per-rnd}
\end{figure}

When a policy is not successful enough, there exists a few common approaches to try and improve the learning of the policy. We take VDN as our baseline since it provides the best results in our experiments and we combine it with Prioritised Experience Replay \citep[PER]{schaul_prioritized_2016_per}, $n$-step return \citep{Watkins_1989_n-step-return} and intrinsic curiosity \citep{jurgen_schmidhuber_possibility_1991_curiosity} on top of it and analyse their impact on the learning process.

\subsubsection{Prioritised Experience Replay} is a technique used in off-policy reinforcement learning to enhance learning efficiency by prioritising experiences and sampling them based on their informativeness. The intuition is to sample past experiences whose $Q$-values are poorly estimated more often and hope that when agents discover a better policy than their current one, this policy would be prioritised. In our setting, we hope that if agents ever complete the level, this experience would be prioritised.

As \citeauthor{schaul_prioritized_2016_per} suggest, we use the temporal difference error as the priority. We have performed a hyperparameter search on $\alpha$ and $\beta$ that respectively control the  exponential scale of priorities and the exponential scale of importance sampling weights with values ranging from 0.3 to 0.8 and have found the best values to be $\alpha=0.6$ and $\beta=0.5$, where $\beta$ is annealed from $0.5$ to $1$ on the course of the training (1m steps).

\autoref{fig:per-rnd} shows that prioritised sampling performs worse than uniform sampling overall. We hypothesise that PER hinders exploration in the early stages of the training because of the zero-incentive dynamics of the game and discuss this further in \autoref{sec:discussion-interdependence} and \autoref{sec:discussion-zid} with regard to interdependence and zero-incentive dynamics respectively. 

\subsubsection{\textit{N}-step return} is a technique that aims at fastening the bootstrapping process \citep{sutton_barto_2018_rlbook} by propagating rewards up to $n$-steps into the past. In the scope of LLE, the idea here is to propagate the reward for collecting gems faster to the step of laser blocking. We have tried values for $n \in \left\{3, 5, 7, 9 \right\}$ and found $n=3$ to give the highest score on average.

\autoref{fig:per-rnd} show that 3-step return yields worse results than any other variant. We relate this poor performance to the high probability of dying from lasers while exploring, resulting in agents learning more conservative policies. We relate this phenomenon to the zero-incentive dynamics of the game and discuss this topic further in \autoref{sec:discussion-zid}. 

\subsubsection{Intrinsic curiosity} aims at introducing bias in the learning process to encourage agents to explore unknown states. During learning, intrinsic curiosity adds an extra reward referred to as the intrinsic reward, thereby altering the updates of the $Q$-function. With this method, we hope that agents would learn faster about laser-blocking and therefore allow them to escape state space bottlenecks.

Intrinsic curiosity with Random Network Distillation \citep[RND]{burda_exploration_2018_rnd} has proven to work well in single-agent environments with zero-incentive dynamics such as Montezuma's Revenge, where the agent must first collect a torch (unrewarded) to be able to light up a dark room (unrewarded) and in the end, collect a treasure (rewarded). Similarly, the objective of using RND is to quicken the discovery of the laser-blocking dynamic and encourage agents to explore the state space.\\
We linearly anneal the intrinsic reward from a factor 2 down to 0 over the course of the training (1m steps), clip the intrinsic reward to be lower or equal to 5 and warm up the RND for 64 updates before issuing any intrinsic reward different from 0.

\autoref{fig:per-rnd} shows that RND performs very similarly to the baseline and does not enable the agents to escape more state space bottlenecks and therefore to learn significantly better policies.

\section{Discussion}
\label{sec:discussion}

\subsection{Perfect coordination}
\label{sec:perfect-coordination}
We analyse the policy that agents learn with the VDN baseline and show their ability to achieve perfect coordination. VDN has been chosen over QMIX for interpretability reasons and because it has been the most successful algorithm in our tests. 

\begin{figure}[t]
    \centering
    \begin{subfigure}[b]{0.24\textwidth}
         \includegraphics[width=\linewidth]{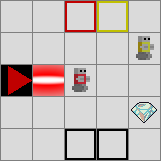}
         \caption{}
         \label{fig:step2-blocking}
    \end{subfigure}
    \begin{subfigure}[b]{0.24\textwidth}
         \includegraphics[width=\linewidth]{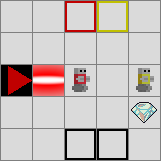}
         \caption{}
         \label{fig:step3-crossing}
    \end{subfigure}
    \begin{subfigure}[b]{0.24\textwidth}
         \includegraphics[width=\linewidth]{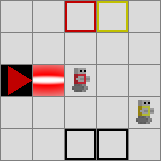}
         \caption{}
         \label{fig:step4-passed}
    \end{subfigure}
    \caption{Four consecutive states of an episode. Agent red blocks the laser for agent yellow and waits for the latter to have left the range of the blocked beam.}
    \label{fig:toy-example}
\end{figure}

\paragraph{Level 6} Analysing the policy of the four agents after training for 1m steps, we can see that the agents understand the laser dynamics: they learn low $Q_i$-values for walking into deadly lasers and for releasing a laser beam on another agent (killing it). Agents red and yellow also learn to cooperate and block lasers for each other in order to reach the bottom half of the map, collect the gems and reach the exit tiles. We argue that agents learn perfect coordination and illustrate that behaviour in a simpler toy example illustrated in \autoref{fig:toy-example}, where agents must also block a laser.

\subsubsection{Toy example} Consider the toy example depicted in \autoref{fig:toy-example} with one laser and two agents. We train the agents for 160k steps and then analyse the $Q_i$-values during the steps concerned by perfect coordination in \autoref{tab:steps-qvalues}.

\begin{table}
    \centering
    \caption{$Q_i$-values from the successive states depicted in \autoref{fig:toy-example}. The highest $Q_i$-values are in bold.}
    \label{tab:steps-qvalues}
    \begin{tabular}{c|l|rrrrr}
         \textbf{State} & \textbf{Agent} & \textbf{North} & \textbf{South} & \textbf{West} & \textbf{East} & \textbf{Stay}\\
         \hline
         \multirow{2}{*}{\ref{fig:step2-blocking}}
         & \cellcolor{red!70} Red & -1.29 & -0.97 & 2.52 & 2.53 & \textbf{2.56} \\
         & \cellcolor{yellow!70} Yellow & 0.78 & \textbf{0.94} & 0.67 & 0.91 & 0.85\\
         \hline
         \multirow{2}{*}{\ref{fig:step3-crossing}}
         & \cellcolor{red!70} Red & 1.44 & 1.51 & 1.48 & 1.51 & \textbf{1.56} \\
         & \cellcolor{yellow!70} Yellow & 1.77 & \textbf{2.13} & 1.18 & 1.41 & 1.45\\
         \hline
         \multirow{2}{*}{\ref{fig:step4-passed}}
         & \cellcolor{red!70} Red & 1.15 & \textbf{1.36} & 1.16 & 1.19 & 1.25 \\
         & \cellcolor{yellow!70} Yellow & 0.71 & 1.42 & \textbf{1.49} & 1.42 & 1.41\\
    \end{tabular}
\end{table}

Starting with step \ref{fig:step2-blocking}, the red agent understands that releasing the laser beam has a much lower value than the other actions because it would likely kill agent yellow. The $Q_i$-values suggest that the \textit{credit} for killing an agent (and hence losing the game) is assigned to the one releasing the beam, not to the one walking in the laser span.
At step \ref{fig:step3-crossing}, the $Q_{i}$-values of agent yellow show that the yellow agent has a clear preference for the south action to collect one gem and close the gap with the exit. Meanwhile, agent red keeps blocking the laser as long as agent yellow stands in the range of the laser beam. With such behaviour similar to the one of level 6 for agent red and yellow, we conclude that agents achieve perfect coordination successfully.

\subsection{Interdependence}
\label{sec:discussion-interdependence}
As discussed in \autoref{sec:prop-interdependence}, agents interdependence introduces bottlenecks in the state space, making that category of environment challenging exploration tasks. We hypothesise that this difficulty of randomly stumbling on good policies contributes to the failure of PER: if agents never come across good policies, PER can never prioritise it. Across the 20 seeded runs with PER on level 6, we checked that at no single point in time, any episode ever reached an exit rate above 0.5. With RND however, agents \emph{very} occasionally reached an exit rate of 0.75, suggesting that RND might be pushed further.

\subsection{Zero-incentive dynamics}
\label{sec:discussion-zid}
We argue that the zero-incentive dynamics of LLE have a detrimental effect on $n$-step return and on PER because they accentuate the punishment of dying from lasers each in their own way.

With regard to PER, since walking into a laser either yields no reward or a punishment (as the definition of zero-incentive dynamics implies), PER is likely to give a higher priority to transitions where punishment has occurred rather than to transitions with a null reward. As a result, we hypothesise that in the early stages of the game, PER emphasises the fact that lasers can be deadly which results in agents being more reluctant to walk into lasers overall.

Looking into the poor performance of $n$-step return, we observe that agents are very likely to die by walking into a laser in their exploration phase. It is also very likely that this punishment is the only non-null reward signal within the $n$ last steps because of the reward sparsity and the zero-incentive dynamics of the environment.  Consequently, when this ``bad'' experience (caused by a poor policy or by random exploration) is then sampled from the replay buffer, the agents learn that the $n$ last actions eventually lead to punishment and their policy is therefore updated to be ``safer''. Our experimental results in Appendix~\ref{apx:n-step} support this explanation, as the mean score decreases when $n$ increases.

\section{Conclusion}
In this paper, we introduced the Laser Learning Environment (LLE), a new cooperative multi-agent grid world that, to the best of our knowledge, exhibits a unique combination of three properties: \textit{perfect coordination}, \textit{interdependence} and \textit{zero-incentive dynamics}. We discussed that the interdependence property induces state space bottlenecks that are difficult to overcome.

We tested IQL, VDN and QMIX against LLE and showed that those algorithms struggled at completing the cooperative task. Our experiments demonstrated that agents successfully achieve perfect coordination but also showed that due to interdependence and zero-incentive dynamics, agents fail at long-term coordination and thus never complete the task. We highlighted that prioritised experience replay and $n$-steps return hinder the agents' performance because of the zero-incentive dynamics of the environment. Intrinsic curiosity with random network distillation also did not provide enough incentive to escape state space bottlenecks and did not enable agents to learn significantly better policies.

Overall, our experiments demonstrated that current state-of-the-art value-based methods fail in LLE and reveal that new benchmarks are needed in cooperative Multi-Agent Reinforcement Learning. This is why we look forward to seeing how the MARL community will approach the Laser Learning Environment and use it to tackle the challenges discussed in this paper and other topics such as generalisation, curriculum learning and inter-agent communication.

\section*{Acknowledgements}
Raphaël Avalos is supported by the FWO (Research Foundation – Flanders) under the grant 11F5721N. Tom Lenaerts is supported by an FWO project (grant nr. G054919N) and two FRS-FNRS PDR (grant numbers 31257234 and 40007793). His is furthermore supported by Service Public de Wallonie Recherche under grant n° 2010235–ariac by digitalwallonia4.ai. Ann Nowé and Tom Lenaerts are also suported by the Flemish Government through the AI Research Program and TAILOR, a project funded by EU Horizon 2020 research and innovation programme under GA No 952215.

\bibliographystyle{plainnat}
\bibliography{biblio}

\begin{thebibliography}{31}
\providecommand{\natexlab}[1]{#1}
\providecommand{\url}[1]{\texttt{#1}}
\expandafter\ifx\csname urlstyle\endcsname\relax
  \providecommand{\doi}[1]{doi: #1}\else
  \providecommand{\doi}{doi: \begingroup \urlstyle{rm}\Url}\fi

\bibitem[Amodei et~al.(2016)Amodei, Olah, Steinhardt, Christiano, Schulman, and
  Man{\'{e}}]{amodei_2016_reward}
Dario Amodei, Chris Olah, Jacob Steinhardt, Paul~F. Christiano, John Schulman,
  and Dan Man{\'{e}}.
\newblock Concrete problems in {AI} safety.
\newblock \emph{CoRR}, abs/1606.06565, 2016.
\newblock URL \url{http://arxiv.org/abs/1606.06565}.

\bibitem[Avalos et~al.(2022)Avalos, Reymond, Nowé, and
  Roijers]{Avalos2022LocalLearningAAMAS}
Raphaël Avalos, Mathieu Reymond, Ann Nowé, and Diederik~M. Roijers.
\newblock {Local Advantage Networks for Cooperative Multi-Agent Reinforcement
  Learning}.
\newblock In \emph{AAMAS '22: Proceedings of the 21st International Conference
  on Autonomous Agents and MultiAgent Systems (Extended Abstract)}, 2022.

\bibitem[Bard et~al.(2020)Bard, Foerster, Chandar, Burch, Lanctot, Song,
  Parisotto, Dumoulin, Moitra, Hughes, Dunning, Mourad, Larochelle, Bellemare,
  and Bowling]{bard_hanabi_2020}
Nolan Bard, Jakob~N. Foerster, Sarath Chandar, Neil Burch, Marc Lanctot,
  H.~Francis Song, Emilio Parisotto, Vincent Dumoulin, Subhodeep Moitra, Edward
  Hughes, Iain Dunning, Shibl Mourad, Hugo Larochelle, Marc~G. Bellemare, and
  Michael Bowling.
\newblock The hanabi challenge: A new frontier for {AI} research.
\newblock \emph{Artificial Intelligence}, 280:\penalty0 103216, 2020.
\newblock ISSN 00043702.
\newblock \doi{10.1016/j.artint.2019.103216}.
\newblock URL \url{http://arxiv.org/abs/1902.00506}.

\bibitem[Boutilier(1996)]{boutilier_planning_1996}
Craig Boutilier.
\newblock Planning, learning and coordination in multiagent decision processes.
\newblock \emph{Proceedings of the 6th Conference on Theoretical Aspects of
  Rationality and Knowledge}, page 195–210, 1996.

\bibitem[Burda et~al.(2018)Burda, Edwards, Storkey, and
  Klimov]{burda_exploration_2018_rnd}
Yuri Burda, Harrison Edwards, Amos Storkey, and Oleg Klimov.
\newblock Exploration by random network distillation, 2018.
\newblock URL \url{http://arxiv.org/abs/1810.12894}.

\bibitem[Cao et~al.(2013)Cao, Yu, Ren, and Chen]{cao_2013_autonomous}
Yongcan Cao, Wenwu Yu, Wei Ren, and Guanrong Chen.
\newblock An overview of recent progress in the study of distributed
  multi-agent coordination.
\newblock \emph{IEEE Transactions on Industrial Informatics}, 9\penalty0
  (1):\penalty0 427--438, 2013.
\newblock \doi{10.1109/TII.2012.2219061}.

\bibitem[Claus and Boutilier(1998)]{claus_dynamics_1998}
Caroline Claus and Craig Boutilier.
\newblock The dynamics of reinforcement learning in cooperative multiagent
  systems.
\newblock \emph{Proceedings of the Fifteenth National Conference on Artificial
  Intelligence and Tenth Innovative Applications of Artificial Intelligence
  Conference, AAAI 98}, 1998.

\bibitem[Klima et~al.(2018)Klima, Bloembergen, Savani, Tuyls, Wittig, Sapera,
  and Izzo]{klima_2018_space}
Richard Klima, Daan Bloembergen, Rahul Savani, Karl Tuyls, Alexander Wittig,
  Andrei Sapera, and Dario Izzo.
\newblock Space debris removal: Learning to cooperate and the price of anarchy.
\newblock \emph{Frontiers in Robotics and AI}, 5, 2018.
\newblock ISSN 2296-9144.
\newblock \doi{10.3389/frobt.2018.00054}.
\newblock URL
  \url{https://www.frontiersin.org/articles/10.3389/frobt.2018.00054}.

\bibitem[Laurent et~al.(2011)Laurent, Matignon, and
  Le~Fort-Piat]{laurent_world_2011_non-markovian}
Guillaume~J. Laurent, Laëtitia Matignon, and N.~Le~Fort-Piat.
\newblock The world of independent learners is not markovian.
\newblock \emph{International Journal of Knowledge-based and Intelligent
  Engineering Systems}, 15\penalty0 (1):\penalty0 55--64, 2011.
\newblock ISSN 18758827, 13272314.
\newblock \doi{10.3233/KES-2010-0206}.
\newblock URL
  \url{https://www.medra.org/servlet/aliasResolver?alias=iospress&doi=10.3233/KES-2010-0206}.

\bibitem[{LeCun} et~al.(1998){LeCun}, Bottou, Bengio, and
  Ha]{lecun_gradient-based_1998_cnn}
Yann {LeCun}, Leon Bottou, Yoshua Bengio, and Patrick Ha.
\newblock Gradient-based learning applied to document recognition, 1998.

\bibitem[Lillicrap et~al.(2016)Lillicrap, Hunt, Pritzel, Heess, Erez, Tassa,
  Silver, and Wierstra]{ddpg_2016}
Timothy~P. Lillicrap, Jonathan~J. Hunt, Alexander Pritzel, Nicolas Heess, Tom
  Erez, Yuval Tassa, David Silver, and Daan Wierstra.
\newblock Continuous control with deep reinforcement learning.
\newblock In Yoshua Bengio and Yann LeCun, editors, \emph{ICLR}, 2016.
\newblock URL
  \url{http://dblp.uni-trier.de/db/conf/iclr/iclr2016.html#LillicrapHPHETS15}.

\bibitem[Lowe et~al.(2017)Lowe, Wu, Tamar, Harb, Abbeel, and
  Mordatch]{mpe_lowe2017multi}
Ryan Lowe, Yi~Wu, Aviv Tamar, Jean Harb, Pieter Abbeel, and Igor Mordatch.
\newblock Multi-agent actor-critic for mixed cooperative-competitive
  environments.
\newblock \emph{Neural Information Processing Systems (NIPS)}, 2017.

\bibitem[Mnih et~al.(2015)Mnih, Kavukcuoglu, Silver, Rusu, Veness, Bellemare,
  Graves, Riedmiller, Fidjeland, Ostrovski, Petersen, Beattie, Sadik,
  Antonoglou, King, Kumaran, Wierstra, Legg, and Hassabis]{dqn_mnih_2015}
Volodymyr Mnih, Koray Kavukcuoglu, David Silver, Andrei~A. Rusu, Joel Veness,
  Marc~G. Bellemare, Alex Graves, Martin Riedmiller, Andreas~K. Fidjeland,
  Georg Ostrovski, Stig Petersen, Charles Beattie, Amir Sadik, Ioannis
  Antonoglou, Helen King, Dharshan Kumaran, Daan Wierstra, Shane Legg, and
  Demis Hassabis.
\newblock Human-level control through deep reinforcement learning.
\newblock \emph{Nature}, 518\penalty0 (7540):\penalty0 529--533, 2015.
\newblock ISSN 1476-4687.
\newblock \doi{10.1038/nature14236}.
\newblock URL \url{https://www.nature.com/articles/nature14236}.
\newblock Number: 7540 Publisher: Nature Publishing Group.

\bibitem[Mordatch and Abbeel(2017)]{mpe_mordatch2017emergence}
Igor Mordatch and Pieter Abbeel.
\newblock Emergence of grounded compositional language in multi-agent
  populations.
\newblock \emph{arXiv preprint arXiv:1703.04908}, 2017.

\bibitem[Oliehoek and Amato(2016)]{oliehoek_concise_2016_pomdp}
Frans~A. Oliehoek and Christopher Amato.
\newblock \emph{A Concise Introduction to Decentralized {POMDPs}}.
\newblock {SpringerBriefs} in Intelligent Systems. Springer International
  Publishing, 2016.
\newblock ISBN 978-3-319-28929-8.
\newblock \doi{10.1007/978-3-319-28929-8}.
\newblock URL \url{http://link.springer.com/10.1007/978-3-319-28929-8}.

\bibitem[Oliehoek et~al.(2008{\natexlab{a}})Oliehoek, Spaan, and
  Vlassis]{oliehoek_optimal_2008_ctde}
Frans~A. Oliehoek, Matthijs T.~J. Spaan, and Nikos Vlassis.
\newblock Optimal and approximate q-value functions for decentralized {POMDPs}.
\newblock \emph{Journal of Artificial Intelligence Research}, 32:\penalty0
  289--353, 2008{\natexlab{a}}.
\newblock ISSN 1076-9757.
\newblock \doi{10.1613/jair.2447}.
\newblock URL \url{http://arxiv.org/abs/1111.0062}.

\bibitem[Oliehoek et~al.(2008{\natexlab{b}})Oliehoek, Spaan, and
  Whiteson]{oliehoek_exploiting_2008_factorisation}
Frans~A Oliehoek, Matthijs T~J Spaan, and Shimon Whiteson.
\newblock Exploiting locality of interaction in factored dec-{POMDPs}.
\newblock \emph{Int. Joint Conf. on Autonomous Agents and Multi-Agent Systems},
  pages 517--524, 2008{\natexlab{b}}.
\newblock URL \url{http://hdl.handle.net/10993/11029}.

\bibitem[Panait and Luke(2005)]{panait_cooperative_2005_marl}
Liviu Panait and Sean Luke.
\newblock Cooperative multi-agent learning: The state of the art.
\newblock \emph{Autonomous Agents and Multi-Agent Systems}, 11\penalty0
  (3):\penalty0 387--434, 2005.
\newblock ISSN 1573-7454.
\newblock \doi{10.1007/s10458-005-2631-2}.
\newblock URL \url{https://doi.org/10.1007/s10458-005-2631-2}.

\bibitem[Parker-Holder et~al.(2022)Parker-Holder, Jiang, Dennis, Samvelyan,
  Foerster, Grefenstette, and Rockt{\"a}schel]{parker2022evolving}
Jack Parker-Holder, Minqi Jiang, Michael Dennis, Mikayel Samvelyan, Jakob
  Foerster, Edward Grefenstette, and Tim Rockt{\"a}schel.
\newblock Evolving curricula with regret-based environment design.
\newblock In \emph{International Conference on Machine Learning}, pages
  17473--17498. PMLR, 2022.

\bibitem[Rashid et~al.(2018)Rashid, Samvelyan, and Schroeder]{rashid_qmix_2018}
Tabish Rashid, Mikayel Samvelyan, and Christian Schroeder.
\newblock {QMIX}: Monotonic value function factorisation for deep multi-agent
  reinforcement learning, 2018.

\bibitem[Samvelyan et~al.(2019)Samvelyan, Rashid, de~Witt, Farquhar, Nardelli,
  Rudner, Hung, Torr, Foerster, and Whiteson]{samvelyan19smac}
Mikayel Samvelyan, Tabish Rashid, Christian~Schroeder de~Witt, Gregory
  Farquhar, Nantas Nardelli, Tim G.~J. Rudner, Chia-Man Hung, Philiph H.~S.
  Torr, Jakob Foerster, and Shimon Whiteson.
\newblock {The} {StarCraft} {Multi}-{Agent} {Challenge}.
\newblock \emph{CoRR}, abs/1902.04043, 2019.

\bibitem[Schaul et~al.(2016)Schaul, Quan, Antonoglou, and
  Silver]{schaul_prioritized_2016_per}
Tom Schaul, John Quan, Ioannis Antonoglou, and David Silver.
\newblock Prioritized experience replay.
\newblock \emph{4th International Conference on Learning Representations,
  {ICLR} 2016 - Conference Track Proceedings}, pages 1--21, 2016.

\bibitem[Schmidhuber(1991)]{jurgen_schmidhuber_possibility_1991_curiosity}
Jürgen Schmidhuber.
\newblock A possibility for implementing curiosity and boredom in
  model-building neural controllers.
\newblock In Jean-Arcady Meyer, editor, \emph{From Animals to Animats}, pages
  222--227. The {MIT} Press, international conference on simulation adaptive
  behavior: from animals to animats edition, 1991.
\newblock ISBN 978-0-262-25667-4.
\newblock \doi{10.7551/mitpress/3115.003.0030}.
\newblock URL
  \url{https://direct.mit.edu/books/book/3865/chapter/162771/a-possibility-for-implementing-curiosity-and}.

\bibitem[Son et~al.(2019)Son, Kim, Kang, Hostallero, and Yi]{son_qtran_2019}
Kyunghwan Son, Daewoo Kim, Wan~Ju Kang, David~Earl Hostallero, and Yung Yi.
\newblock {QTRAN}: Learning to factorize with transformation for cooperative
  multi-agent reinforcement learning, 2019.
\newblock URL \url{http://arxiv.org/abs/1905.05408}.

\bibitem[Stern et~al.(2017)Stern, Sturtevant, Felner, Koenig, Ma, Walker, Li,
  Atzmon, Cohen, Kumar, Boyarski, and Bart]{stern_multi-agent_2017_conflicts}
Roni Stern, Nathan~R Sturtevant, Ariel Felner, Sven Koenig, Hang Ma, Thayne~T
  Walker, Jiaoyang Li, Dor Atzmon, Liron Cohen, T~K~Satish Kumar, Eli Boyarski,
  and Roman Bart.
\newblock Multi-agent pathfinding: Definitions, variants, and benchmarks, 2017.

\bibitem[Sunehag et~al.(2018)Sunehag, Lever, Gruslys, Czarnecki, Zambaldi,
  Jaderberg, Lanctot, Sonnerat, Leibo, Tuyls, and
  Graepel]{vdn_sunehag_value-decomposition_2018}
Peter Sunehag, Guy Lever, Audrunas Gruslys, Wojciech~Marian Czarnecki, Vinicius
  Zambaldi, Max Jaderberg, Marc Lanctot, Nicolas Sonnerat, Joel~Z. Leibo, Karl
  Tuyls, and Thore Graepel.
\newblock Value-decomposition networks for cooperative multi-agent learning
  based on team reward.
\newblock \emph{Proceedings of the International Joint Conference on Autonomous
  Agents and Multiagent Systems, {AAMAS}}, 3:\penalty0 2085--2087, 2018.
\newblock ISSN 15582914.
\newblock {ISBN}: 9781510868083.

\bibitem[Sutton and Barto(2018)]{sutton_barto_2018_rlbook}
Richard~S. Sutton and Andrew~G. Barto.
\newblock Reinforcement learning: an introduction, 2018.

\bibitem[Sutton(1984)]{sutton_1984_credit_assignment}
Richard~Stuart Sutton.
\newblock \emph{Temporal Credit Assignment in Reinforcement Learning}.
\newblock PhD thesis, University of Massachusetts Amherst, 1984.
\newblock AAI8410337.

\bibitem[Tuyls and Weiss(2012)]{tuyls_multiagent_2012_moving_target}
Karl Tuyls and Gerhard Weiss.
\newblock Multiagent learning: Basics, challenges, and prospects.
\newblock \emph{{AI} Magazine}, 33\penalty0 (3):\penalty0 41, 2012.
\newblock ISSN 2371-9621, 0738-4602.
\newblock \doi{10.1609/aimag.v33i3.2426}.
\newblock URL
  \url{https://ojs.aaai.org/index.php/aimagazine/article/view/2426}.

\bibitem[Watkins(1989)]{Watkins_1989_n-step-return}
Christopher Watkins.
\newblock \emph{Learning From Delayed Rewards}.
\newblock PhD thesis, University of Cambridge, 1989.

\bibitem[Wu et~al.(2021)Wu, Wang, Evans, Tenenbaum, Parkes, and
  Kleiman-Weiner]{overcooked_wu_wang2021too}
Sarah~A. Wu, Rose~E. Wang, James~A. Evans, Joshua~B. Tenenbaum, David~C.
  Parkes, and Max Kleiman-Weiner.
\newblock Too many cooks: Coordinating multi-agent collaboration through
  inverse planning.
\newblock \emph{Topics in Cognitive Science}, 2021.
\newblock \doi{https://doi.org/10.1111/tops.12525}.
\newblock URL \url{https://onlinelibrary.wiley.com/doi/abs/10.1111/tops.12525}.

\end{thebibliography}
\newpage

\appendix
\section{Code repository}
All the code for the Laser Learning Environment is publicly available on the repository \url{https://github.com/yamoling/lle}.

\section{Hyperparameters}
\label{apx:hyperparameters}
Hyperparameter search was performed with VDN on a combination of batch sizes (32, 64 and 128 transitions) and memory sizes (50k, 100k and 200k transitions). Then, we have tried training intervals of 1 and 5.

Then, we performed a hyperparameter search for prioritised experience replay on a combination of $\alpha$ (0.3, 0.4, 0.5, 0.6, 0.7, 0.8) and $\beta$ (0.3, 0.4, 0.5, 0.6, 0.7, 0.8).

For random network distillation, we have explored update ratios $p = 0.25, 0.5$ and $1$, then enabled or disabled annealing, and finally tried clipping the intrinsic reward to 1 or to 5.

\begin{table}[h]
    \centering
    \caption{Hyperparameters used across all the experiments}
    \begin{tabular}{l|l|p{7cm}}
        \textbf{Parameter} & \textbf{Value} & \textbf{Comment} \\
        \hline
        Memory size & 50 000 & Transitions\\
        Batch size & 64 & Transitions\\
        Train interval & 5 time steps & \\
        Optimiser & Adam & \\
        $\alpha$ (Learning rate) & 0.0005 & For both $Q$-network and mixer \\
        Grad norm clipping & 10 & Clips both $Q$-network and mixer\\
        $\gamma$ & $0.95$ & Discount factor\\
        $\tau$ & $0.01$ & Soft update rate\\
        $\epsilon_{start}$ & $1$ &\\
        $\epsilon_{min}$ & $0.05$ &\\
        $\epsilon$ annealing & 500k time steps & Linearly annealed\\
        \hline
        $\alpha$ & 0.6 & PER \\
        $\beta$ & 0.5 & PER \\
        $\beta$ annealing & 1m time steps & PER\\
        \hline
        $p$ & $0.25$ & Randomly mask error from RND with probability $p$\\
        IR factor & $2 \rightarrow 0$ & Intrinsic Reward linearly annealed over 1m steps\\
        IR clip & $5$ & IR is clipped to 5 maximum\\
        IR warmup & 64 & The RND is optimised 64 times before issuing any intrinsic reward\\
        \hline
    \end{tabular}
\end{table}

\newpage
\section{Neural network architecture}
\label{apx:nn_architecture}
The $Q$-network is made of two parts with an interconnection: a convolutional neural network of three layers, an interconnect that flattens the CNN, and finally a neural network of three linear layers. This is depicted in \autoref{tab:nn}.

\begin{table}[h]
    \centering
    \caption{$Q$-network architecture}
    \label{tab:nn}
    \begin{tabular}{|llrcr|}
        \hline
         \textbf{Layer type} & \textbf{Activation} & \textbf{Output shape} & \textbf{Stride} & \textbf{Kernel} \\
         \hline
         Input & & $11 \times 13 \times 15$ & &\\
         Conv2D & ReLU & $32 \times 11 \times 13$ & $1$ & $(3 \times 3)$\\
         Conv2D & ReLU & $64 \times 9 \times 11$ & $1$ & $(3 \times 3)$\\
         Conv2D & ReLU & $32 \times 7 \times 9$ & $1$ &  $(3 \times 3)$\\
         Flatten & & $2016$ &&\\
         Concat & & $2020$&&\\
         Linear & ReLU & $64$ &&\\
         Linear & ReLU & $64$ &&\\
         Linear & & $5$ &&\\
         \hline
    \end{tabular}
\end{table}

\section{Results of \textit{n}-steps returns with VDN}
\label{apx:n-step}

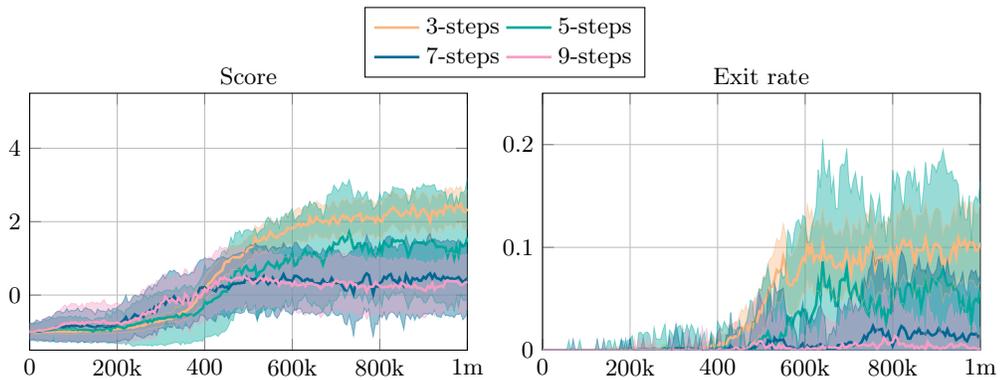
\begin{figure}[h]
    \centering
    \begin{tikzpicture}
        \begin{groupplot}[
            group style={group size= 2 by 1},
            height=5cm,
            width=7.4cm, 
            grid=major,
            legend columns=2
            xtick={0, 200000, 400000, 600000, 800000, 1000000},
            xticklabels={0, 0, 200k, 400, 600k, 800k, 1m},
            scaled x ticks=false,
            title style={yshift=-0.2cm},
        ]
            \nextgroupplot[legend to name=luigi, title={Score}, ymin=-1.5, ymax=5.5, xmin=0, xmax=1000000]       
                \plotMean[3-steps]{3step}{score};
                \plotMean[5-steps]{5step}{score};
                \plotMean[7-steps]{7step}{score};
                \plotMean[9-steps]{9step}{score};
            
                \plotCI{3step}{score};
                \plotStd{5step}{score};
                \plotStd{7step}{score};
                \plotStd{9step}{score};

                \coordinate (top-left) at (rel axis cs:0,1);

            \nextgroupplot[title={Exit rate}, ymin=0, ymax=0.25, xmin=0, xmax=1000000]       
                \plotMean{3step}{exit_rate};
                \plotMean{5step}{exit_rate};
                \plotMean{7step}{exit_rate};
                \plotMean{9step}{exit_rate};
            
                \plotCI{3step}{exit_rate};
                \plotStd{5step}{exit_rate};
                \plotStd{7step}{exit_rate};
                \plotStd{9step}{exit_rate};

                \coordinate (top-right) at (rel axis cs:1,1);
        \end{groupplot}
        \path (top-left)--(top-right) coordinate[midway] (h-center);
        \node[above=0.2cm, inner sep=0pt] at (h-center) {\pgfplotslegendfromname{luigi}};
    \end{tikzpicture}
    \caption{Scores and exit rates for different values of $n$ in $n$-steps return. Results are shown with the mean in bold $\pm$ standard deviation, capped by minimum and maximum.}
\end{figure}

\newpage
\section{Maps provided by LLE}
\label{apx:maps}

\begin{figure}[h]
    \centering
    \begin{subfigure}[t]{0.31\linewidth}
        \includegraphics[width=\linewidth]{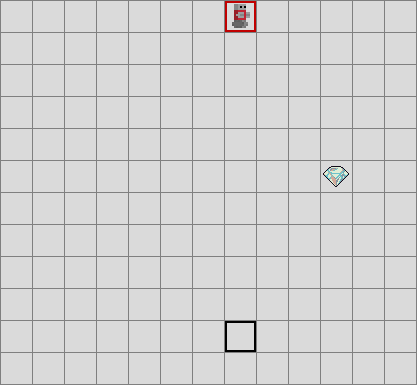}
        \caption{Level 1 for debugging purposes.\newline}
    \end{subfigure}
    \hfill
    \begin{subfigure}[t]{0.31\linewidth}
        \includegraphics[width=\linewidth]{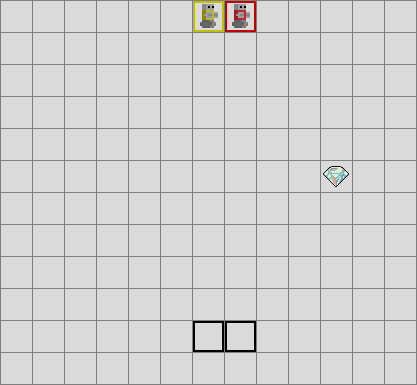}
        \caption{Level 2, first step into multi-agent problems, almost no interdependence.}
    \end{subfigure}
    \hfill
    \begin{subfigure}[t]{0.31\linewidth}
        \includegraphics[width=\linewidth]{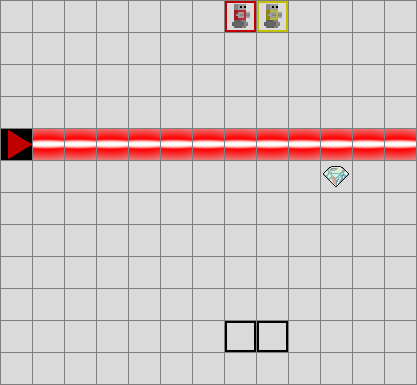}
        \caption{Level 3, introduces the dynamic of laser-blocking, which increases interdependence.}
    \end{subfigure}
    \begin{subfigure}[t]{0.31\linewidth}
        \includegraphics[width=\linewidth]{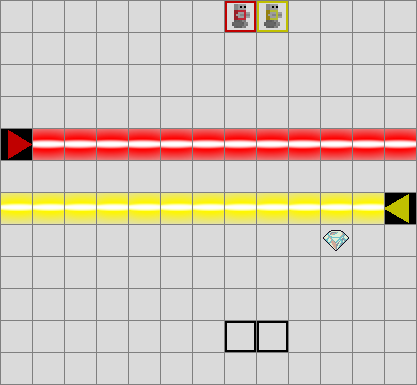}
        \caption{Level 4 introduces two lasers such that each agent successively has to block it for the other for even more interdependence.}
    \end{subfigure}
    \hfill
    \begin{subfigure}[t]{0.31\linewidth}
        \includegraphics[width=\linewidth]{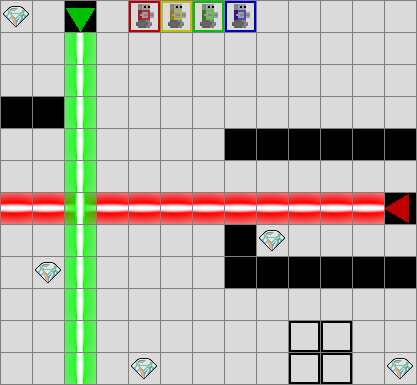}
        \caption{Level 5 also has 2 lasers but has 4 agents, which increases interdependence.}
    \end{subfigure}
    \hfill
    \begin{subfigure}[t]{0.31\linewidth}
        \includegraphics[width=\linewidth]{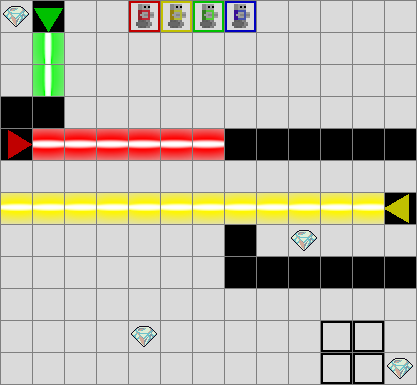}
        \caption{Level 6, four agents and three lasers, the most difficult level introduced with the highest level of interdependence.}
    \end{subfigure}
    \caption{Six standard levels of LLE. The levels have been designed with incremental level of interdependence in mind. Level 6 is the main level studied in this work.}
    \label{fig:my_label}
\end{figure}

\newpage
\section{Results on standard maps}
\label{apx:all-results}

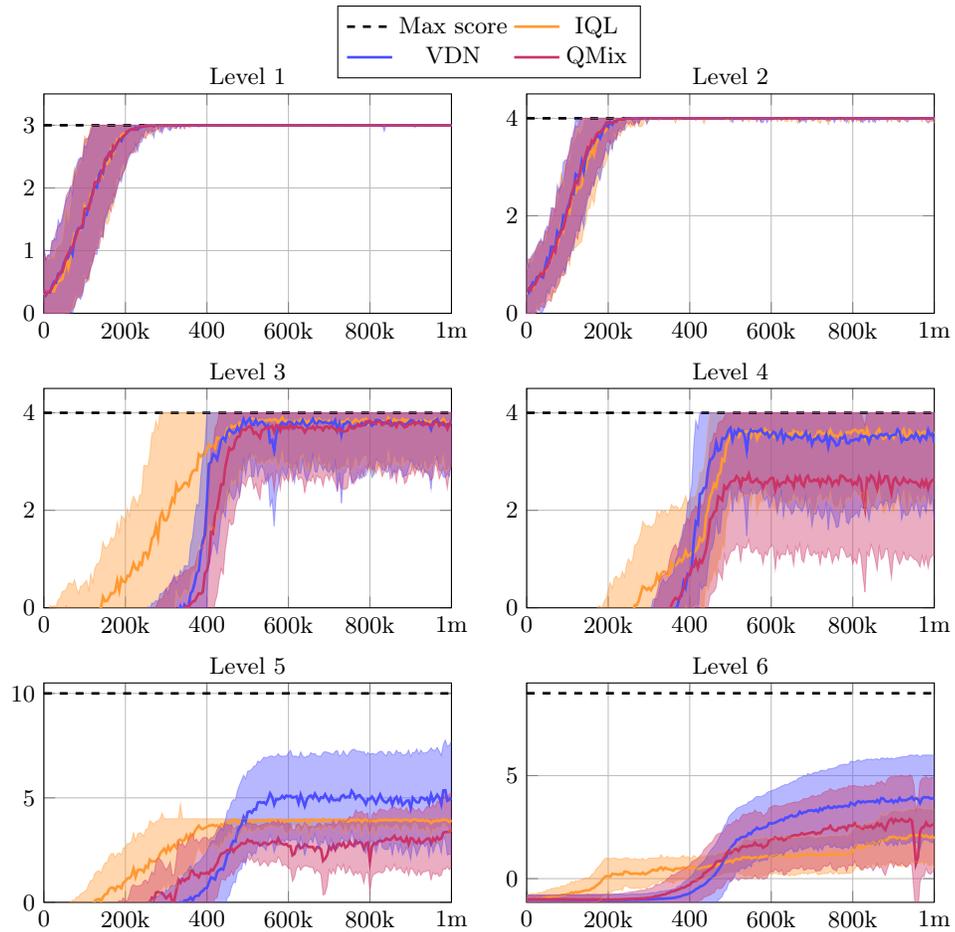
\begin{figure}[h]
    \centering
    \begin{tikzpicture}
        \begin{groupplot}[
            group style={group size= 2 by 3},
            height=4.5cm,
            width=7cm, 
            grid=major,
            legend columns=2
            xtick={0, 200000, 400000, 600000, 800000, 1000000},
            xticklabels={0, 0, 200k, 400, 600k, 800k, 1m},
            scaled x ticks=false,
            title style={yshift=-0.2cm},
        ]
            \nextgroupplot[title={Level 1}, legend to name=mario, ymin=0, ymax=3.5, xmin=0, xmax=1000000]       
                \addplot[line width=1pt, mark=none, color=black, style=dashed] coordinates {(0,3) (1000000,3)};
                \addlegendentry{Max score}

                \plotLevelMean[IQL]{dqn}{lvl1};
                \plotLevelMean[VDN]{vdn}{lvl1};
                \plotLevelMean[QMix]{qmix}{lvl1};
            
                \plotLevelStd{dqn}{lvl1};
                \plotLevelStd{vdn}{lvl1};
                \plotLevelStd{qmix}{lvl1};

                \coordinate (top-left) at (rel axis cs:0,1);
                \coordinate (bot-left) at (rel axis cs:0,0);
                
            \nextgroupplot[title={Level 2}, ymin=0, ymax=4.5, xmin=0, xmax=1000000]
                \addplot[line width=1pt, mark=none, color=black, style=dashed] coordinates {(0,4) (1000000,4)};
                \plotLevelMean{dqn}{lvl2};
                \plotLevelMean{vdn}{lvl2};
                \plotLevelMean{qmix}{lvl2};
                \addplot[black, samples=2] {0.5};
            
                \plotLevelStd{dqn}{lvl2};
                \plotLevelStd{vdn}{lvl2};
                \plotLevelStd{qmix}{lvl2};
                \coordinate (right) at (rel axis cs:1,1);
                \coordinate (bot) at (rel axis cs:1,0);

            \nextgroupplot[title={Level 3}, ymin=0, ymax=4.5, xmin=0, xmax=1000000]
                \addplot[line width=1pt, mark=none, color=black, style=dashed] coordinates {(0,4) (1000000,4)};
                \plotLevelMean{dqn}{lvl3};
                \plotLevelMean{vdn}{lvl3};
                \plotLevelMean{qmix}{lvl3};
            
                \plotLevelStd{dqn}{lvl3};
                \plotLevelStd{vdn}{lvl3};
                \plotLevelStd{qmix}{lvl3};

            \nextgroupplot[title={Level 4}, ymin=0, ymax=4.5, xmin=0, xmax=1000000]
                \addplot[line width=1pt, mark=none, color=black, style=dashed] coordinates {(0,4) (1000000,4)};
                \plotLevelMean{dqn}{lvl4};
                \plotLevelMean{vdn}{lvl4};
                \plotLevelMean{qmix}{lvl4};
            
                \plotLevelStd{dqn}{lvl4};
                \plotLevelStd{vdn}{lvl4};
                \plotLevelStd{qmix}{lvl4};

            \nextgroupplot[title={Level 5}, ymin=0, ymax=10.5, xmin=0, xmax=1000000]
                \addplot[line width=1pt, mark=none, color=black, style=dashed] coordinates {(0,10) (1000000,10)};
                \plotLevelMean{dqn}{lvl5};
                \plotLevelMean{vdn}{lvl5};
                \plotLevelMean{qmix}{lvl5};
            
                \plotLevelStd{dqn}{lvl5};
                \plotLevelStd{vdn}{lvl5};
                \plotLevelStd{qmix}{lvl5};

            \nextgroupplot[title={Level 6}, ymin=-1.15, ymax=9.5, xmin=0, xmax=1000000]
                \addplot[line width=1pt, mark=none, color=black, style=dashed] coordinates {(0,9) (1000000,9)};
                \plotMean{dqn}{score};
                \plotMean{vdn}{score};
                \plotMean{qmix}{score};
            
                \plotStd{dqn}{score};
                \plotStd{vdn}{score};            
                \plotStd{qmix}{score};

        \end{groupplot}
        \path (top-left)--(bot) coordinate[midway] (center);
        \path (top-left)--(bot-left) coordinate[midway] (center-left);
        \path (top-left)--(right) coordinate[midway] (h-center);
        \node[above=0.2cm, inner sep=0pt] at (h-center) {\pgfplotslegendfromname{mario}};
    \end{tikzpicture}
    \caption{Scores on standard maps over training time. Maximum score achievable is shown as a black dotted line. These results show the mean in bold $\pm$ the standard deviation, capped by minimum and maximum.}
\end{figure}

\end{document}